\documentclass[journal]{IEEEtran}
\usepackage[left=51pt,top=58pt,right=52pt,bottom=47pt]{geometry}
\usepackage{graphicx} 

\usepackage[margin=0pt,font=small,labelfont=bf]{caption}

\usepackage{amsmath,amsthm,amssymb,amsfonts}
\usepackage{bm}

\usepackage{pifont}
\usepackage{stfloats}
\usepackage{subcaption}
\usepackage{lipsum}  

\usepackage[numbers,sort&compress,square]{natbib}

\usepackage{subcaption}
\usepackage{rotating, graphicx}
\usepackage{tabularx}
\usepackage{algorithmic}
\usepackage[titlenumbered,ruled]{algorithm2e}

\usepackage{booktabs}
\usepackage{multirow}
\usepackage{booktabs} 
\usepackage[colorlinks,bookmarksopen,bookmarksnumbered,linkcolor=blue,citecolor=blue,urlcolor=blue]{hyperref}

\newcommand{\etal}{\textit{et al.}}

\begin{document}
%
\title{Contact-Prioritized Planning of Impact-Resilient Aerial Robots with an Integrated Compliant Arm}

\author{Zhichao~Liu,$^1$ 
        Zhouyu~Lu,$^1$ 
        Ali-akbar~Agha-mohammadi,$^2$ 
        and~Konstantinos~Karydis$^1$

\thanks{$^1$ Dept. of Electrical and Computer Engineering, University of California, Riverside 
	(email: {\{zliu157, zlu044,  karydis\}@ucr.edu}).} 

\thanks{$^2$ Field AI (email: ali@fieldai.com).}
 
\thanks{We gratefully acknowledge the support of NSF grants \#IIS-1724341, \#IIS-1901379 and \#IIS-1910087, ARL grant \#W911NF-18-1-0266, and ONR grant \#N00014-19-1-2264. Any opinions, findings, and conclusions or recommendations expressed in this material are those of the authors and do not necessarily reflect the views of the funding agencies.}%
}

%
%

\markboth{}%
{Liu \MakeLowercase{\textit{et al.}}: Contact-Prioritized Planning of Impact-Resilient Aerial Robots with an Integrated Compliant Arm}
%



\maketitle

\begin{abstract}
The article develops an impact-resilient aerial robot (s-ARQ) equipped with a compliant arm to sense contacts and reduce collision impact and featuring a real-time contact force estimator and a non-linear motion controller to handle collisions while performing aggressive maneuvers and stabilize from high-speed wall collisions. Further, a new collision-inclusive planning method that aims to prioritize contacts to facilitate aerial robot navigation in cluttered environments is proposed. A range of simulated and physical experiments demonstrate key benefits of the robot and the contact-prioritized (CP) planner.
Experimental results show that the compliant robot has only a $4\%$ weight increase but around $40\%$ impact reduction in drop tests and wall collision tests. 
s-ARQ can handle collisions while performing aggressive maneuvers and stabilize from high-speed wall collisions at $3.0$ m/s with a success rate of $100\%$. 
Our proposed compliant robot and contact-prioritized planning method can accelerate computation time while having shorter trajectory time and larger clearances compared to A$^\ast$ and RRT$^\ast$ planners with velocity constraints. Online planning tests in partially-known environments further demonstrate the preliminary feasibility of our method to apply in practical use cases. 
\end{abstract}


%
\IEEEpeerreviewmaketitle

\section{Introduction}
%
%
%
%
\IEEEPARstart{M}{icro} Aerial Vehicles (MAVs) can support sensor-based exploration and navigation, and en route to robust autonomous navigation, aerial autonomy with interactive behavior has been studied~\cite{bartelds2016compliant, kim2016vision, park2018odar}. There has been a growing interest in deploying MAVs in challenging environments, including but not limited to confined~\cite{de2021resilient, lew2019contact} and cluttered~\cite{mulgaonkar2017robust} ones. Collision risks get significantly higher for autonomous missions in these complex environments. 
Compliant resilient robots attract growing attention due to the merits of reducing impact and protecting onboard sensors~\cite{Patnaik}. Taking advantage of impact resilience, research efforts on collision-inclusive motion planning have started to be proposed~\cite{zha2021exploiting, zlu2021iros, lu2022online}. 

In this work, we introduce a lightweight compliant arm to sense contacts and reduce high-speed collision impact. Equipped with the integrated compliant arm, we develop an impact-resilient aerial robot (named s-ARQ). The compliant robot has only a $4\%$ weight increase compared to its rigid counterpart, however experimental results show that the compliant arm can reduce around $40\%$ impact. The compliant arm incorporates a passive spring and a laser ranging sensor to enable contact force estimation. Employing a force estimator and non-linear motion controller, s-ARQ can stabilize from high-speed wall collisions at $3.0$ m/s with a success rate of $100\%$. We consolidate the impact resilience by including pole obstacle collisions, as well as different yaw angles. Further, we harness s-ARQ’s strong collision resilience capability to propose a novel planning method that prioritizes contacts. Physical tests and extended simulations demonstrate that our proposed compliant robot and contact-prioritized (CP) planner can accelerate computation while achieving shorter trajectory time and larger clearances compared to collision-avoidance methods with velocity constraints. 
Online planning tests in partially-known environments were studied to support application toward practical use cases. Simulated results further validate the efficiency of the proposed CP planner.

\section{Related Works}
\subsection{Impact-Resilient Compliant MAVs}
Several compliant aerial robots have been developed over the years. Bartelds \etal~\cite{bartelds2016compliant} developed an aerial robot with a compliant contact arm. Examples of works integrating compliant protective structure onto the robot include origami-inspired mechanisms~\cite{sareh2018rotorigami, shu2019quadrotor} and an icosahedron tensegrity structure~\cite{zha2020collision}. De Petris \etal~\cite{de2021resilient} studied impact-resilient MAVs with external compliant flaps. Compliance has also been included into the robot chassis to reduce impact. Chen \etal~\cite{chen2021collision} developed a collision-resilient insect-scale compliant flapping-wing robot, while Patnaik \etal~\cite{Patnaik} introduced a collision-resilient MAV with foldable arms. Soft aerial robots were developed for physical interactions~\cite{nguyen2022soft, ruiz2022sophie}. Further, insect-inspired multicopters are presented with compliant frames to handle collisions~\cite{mintchev2017insect} whereas compliant frames based on tensegrity~\cite{savin2022mixed} are proved helpful to reduce impact. 
Our prior work developed a collision-resilient MAV with compliant arms~\cite{liu2021arq}. Compliant MAVs can reduce the effect of impact and help survive collisions; however, these robots cannot estimate contact force and handle impacts accordingly. 

\vspace{-6pt}
\subsection{Contact Force Estimation}
This work focuses on estimating contact force (excluding external torque). One way is to map robot control inputs to external contact forces offline~\cite{bellens2012hybrid}. However, this applies only when the robot is in contact. In a different approach, momentum-based external wrench estimators with second-order estimation dynamics~\cite{ tomic2017external, lew2019contact} and a Lyapunov-based nonlinear external wrench observer including also inertia shaping~\cite{yuksel2014nonlinear} have been proposed. Unscented Kalman filters are also utilized for estimation~\cite{ mckinnon2016unscented}. A nonlinear disturbance observer has been proposed to estimate contact force~\cite{fan2022quadrotor}. Recently, the feasibility of using cameras to estimate contact force was shown~\cite{stephens2022integrated}. Yet, external force estimation in presence of compliant frames is an open task for aerial robots.  

\vspace{-6pt}
\subsection{Physical Interaction}
Aerial robots are equipped with end-effectors to physically interact with environments. Bartelds \etal~\cite{bartelds2016compliant} studied a compliant manipulator for impact reduction. Flying robots with end-effectors are utilized to apply a force to vertical walls~\cite{wopereis2017application, hamaza2018adaptive}, inspection~\cite{praveen2020inspection} and sensor placement~\cite{stephens2022integrated}. Alejandro \etal~\cite{suarez2021cartesian} developed a compliant arm to monitor and control interaction wrench. 
However, these projects focus on low-speed interactions, while high-speed collisions involve large impact forces and attitude changes. The interactions fail to assist with motion tasks such as planning and exploration.

\vspace{-6pt}
\subsection{Contact-Inclusive Planning}
Contrary to collision avoidance, impact-resilient robots can embrace contacts to improve overall safety and navigation task effectiveness. Mote \etal~\cite{mote2020collision} incorporated the collision model into mixed integer programming for trajectory optimization. Contacts can be also used to improve velocity estimation~\cite{lew2019contact} and mapping~\cite{mulgaonkar2020tiercel}. Risk reward trade-offs have been studied for collision-inclusive trajectories~\cite{lu2020motion, de2022risk}. Local re-planners with setpoint adjustment post collision can be adopted to improve global planners like A$^\star$~\cite{zlu2021iros, lu2022online} and sampling methods~\cite{zha2021exploiting}. However, these methods directly extend global planners and can bound to their constraints.

\section{Development and Key Features of S-ARQ}
\subsection{Design}
\begin{figure}[!t]
\begin{center}
\includegraphics[width=0.4\textwidth]{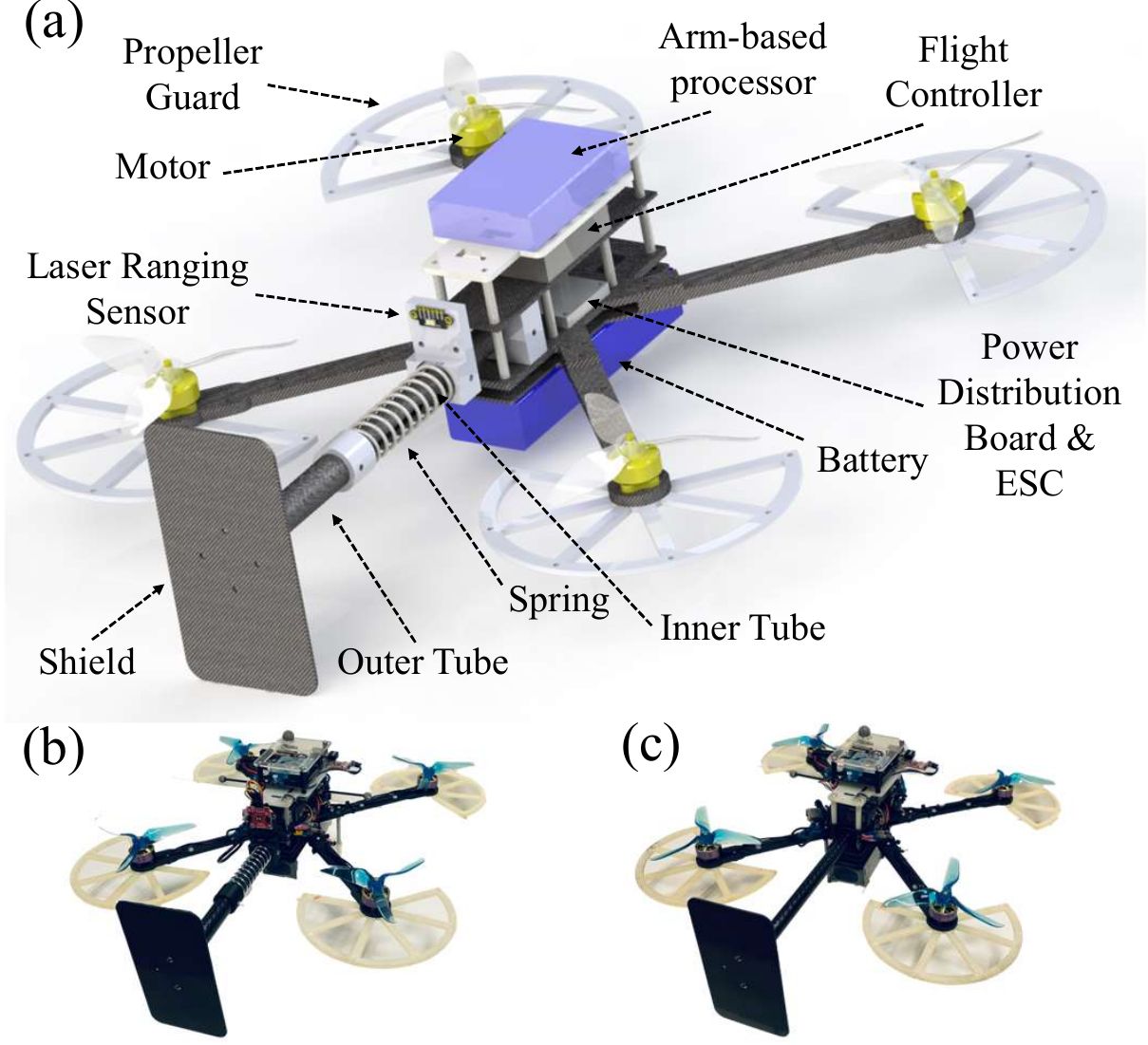}
\end{center}
\vspace{-6pt}
\caption{The impact-resilient aerial robot introduced in this work. (a) CAD rendering of the robot. Physical prototypes of the (b) compliant robot and (c) its rigid counterpart.}
\label{fig:design}
\vspace{-12pt}
\end{figure}

Inspired by prior works on compliant end-effectors~\cite{bartelds2016compliant}, we introduce a lightweight compliant arm design and embed it onto a custom-made quadrotor to enable the latter to both estimate contact forces and stabilize from high-speed collisions.  
The robot introduced in this work is named \underline{s}ingle-arm \underline{A}ctive \underline{R}esilient \underline{Q}uadrotor (s-ARQ). When moving forward in static environments (the robot's front faces the moving direction), robots mostly have contacts only in the front direction. We revise our earlier compliant aerial robot design~\cite{liu2021arq} to attach one compliant arm onto the (rigid otherwise) chassis (Fig.~\ref{fig:design}). 

The new arm design consists of two carbon fiber tubes, a compression steel spring, a laser range sensor, and a carbon fiber shield. Carbon fiber tubes (tensile strength $125,000$-$175,000$ psi) are assembled in a concentric manner. The outer tube has an outer diameter (OD) of $14.5$ mm while the inner tube has an OD of $17.5$ mm. The inner tube includes a linear slot to limit rotational motion with negligible friction, thus both tubes comprise a prismatic joint. A steel compression spring (OD $21.5$ mm, free length $76$ mm) connects both tubes. A lightweight fiber sheet ($75\times140$ mm) is fabricated with a Stepcraft D.600 CNC router with enclosure and milling bath, to work as the end-effector (shield) to contact with obstacles. Multiple custom adapters are 3D-printed with a Markforged Mark 2 printer. A laser range sensor (VL53L1X) is attached to the inner tube to measure the length of the compliant arm. Our design differs from works~\cite{bartelds2016compliant, stephens2022integrated} in its capacity to enable stabilization from high-speed collisions with large impact, and estimate contact forces in presence of frame compliance. 
 
The robot chassis is shared with our prior work~\cite{liu2022safely} and consists of custom carbon fiber frames, a flight controller (Pixhawk), and an ARM-based multi-core processor (Odroid). The four frame arms measure $0.19$ m, and the contact arm measures $0.28$ m in free flight. 
The compliant robot weighs $890$ g without batteries. For comparison purposes, we also build a rigid version (Quad), which shares the same quadrotor platform but a rigid contact arm of the same length (Fig.~\ref{fig:design}(c)). S-ARQ is $50$ g heavier ($4\%$ of the total weight with batteries) compared to its rigid counterpart. We use a $5200$ mAh Lipo battery yielding a flight time of approximately $610$ sec.       

\vspace{6pt}
\vspace{-10pt}

\begin{figure}[!h]
\begin{center}
\includegraphics[width=0.25\textwidth]{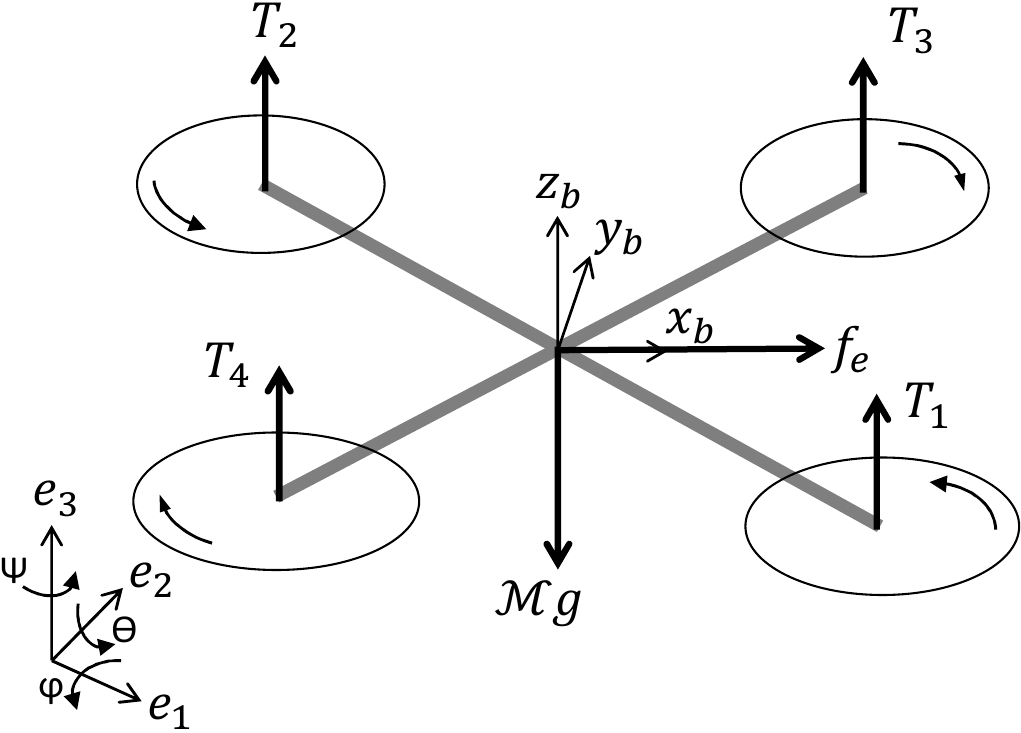}
\end{center}
\vspace{-6pt}
\caption{Dynamic model of the system.}
\label{fig:model}
\vspace{-12pt}
\end{figure}

\vspace{-12pt}
\subsection{Modeling}
With reference to Fig.~\ref{fig:model}, the equations of motion for our robot following notation~\cite{tomic2017external} are
\begin{equation}\label{Eqn:dynamic}
		\begin{aligned}
		\mathcal{M}\Ddot{\bm{r}}  &= -\mathcal{M}g \bm{e}_3 + \bm{R} \bm{f}_T + \bm{R}\bm{f}_e  \\
		 \mathcal{I}\Dot{\bm{\omega}} &= \mathcal{S}(\bm{\mathcal{I}}\bm{\omega})\bm{\omega} + \bm{m}_T + \bm{m}_e \\
			\Dot{\bm{R}} &= \bm{R} \mathcal{S}(\bm{\omega})
		\end{aligned}
\end{equation}
where $\bm{r} = [x\ y\ z]^T$ is the position in the inertial frame (East-North-Up), $\mathcal{M}$ is the mass, and $\bm{R} \in SO(3)$ denotes the rotation matrix from body (Forward-Left-Up) to inertial frame. $\mathcal{S}(\cdot)$ is the skew-symmetric operator, $g = 9.81\ \text{m/s}^2$ is the gravity constant, $\bm{e}_3 = [0\ 0\ 1]^T$, $\bm{f}_T$ and $\bm{f}_e = [f_e\ 0\ 0]^T$ are the thrust and external force vectors in body frame, respectively, and $\bm{m}_T$ and $\bm{m}_e$ are the moments generated by the thrust and external force vectors, respectively. Note that, as in~\cite{fumagalli2014developing}, the external moment ($\bm{m}_e$) is not considered herein.


\vspace{-6pt}
\subsection{Contact Force Estimation} \label{sec:est} 
The contact force along the compliant arm can be measured utilizing Hooke's law as
\begin{equation*}
    \hat{f}_e = -k_l (\delta l + l_0)\enspace,
\end{equation*}
where $\delta l = l_{\text{max}} - \hat{l}$ is the arm length difference ($l_{\text{max}} = 0.28$ m) and $\hat{l}$ is the estimated arm length measured by the onboard distance sensor. To prevent oscillations, we pre-load the compliant arm ($l_0 = 2$ mm). The selected spring constant is $k_l=3.80$ N/mm. The distance sensor has precision of $1$ mm with accuracy of $\pm5$ mm. To mitigate sensor noise we apply a  recursive filter ($w = 0.6$) to sensor readings ($h_i$) as
\begin{equation}\label{eq:filter}
    \hat{l}_i = w h_i + (1-w) \hat{l}_{i-1}\enspace.
\end{equation} 

The distance sensor has a frequency of around $25$ Hz. Admittedly, this design can only estimate forces along the contact arm ($\bm{x}_b$ axis) in body frame. When flying toward known obstacles, the robot can face obstacles along its $\bm{x}_b$ axis utilizing yaw control. Although the compliant arm is of no help to estimate external torques, the prismatic joint does not affect estimating methods such as~\cite{yuksel2014nonlinear, tomic2017external}.

\section{Motion Control and Collision Handling}

To stabilize after high-speed collision, the tracking controller of the robot must be able to follow aggressive trajectories with large attitude angles. In this work, we adopt the cascaded tracking control method as in our prior work~\cite{liu2021arq}.  
Note that the attitude is described as roll ($\phi$), pitch ($\theta$) and yaw ($\psi$) angles, such that $\bm{R} = \bm{R}_z(\psi)\bm{R}_y(\theta)\bm{R}_x(\phi)$, where $\bm{R}_x, \bm{R}_y, \bm{R}_z$ are elementary rotation matrices about the coordinate axis (see Fig.~\ref{fig:model}). The planner generates desired states (position $\bm{r}_\text{des}$, velocity $\Dot{\bm{r}}_\text{des}$ acceleration $\Ddot{\bm{r}}_\text{des}$ and yaw angle $\psi_\text{des}$). The tracking controller comprises high-level position control, mid-level attitude and bodyrate control, and a low-level mixer to output PWM signals to actuators.   

The position controller harnesses geometric constraints for nonlinear tracking~\cite{lee2010geometric, mellinger2011minimum}. First we find the desired thrust force vector $\bm{F}_\text{des} \in \mathbb{R}^3$ in the inertial frame
\vspace{-1pt}
\begin{equation*}\label{eq:f_t_inerial}
\bm{F}_\text{des} = -\bm{K}_d(\Dot{\bm{r}} - \Dot{\bm{r}}_\text{des}) - \bm{K}_p(\bm{r} - \bm{r}_\text{des}) + \mathcal{M}\Ddot{\bm{r}}_\text{des} + \mathcal{M}g\bm{e}_3\;,
\end{equation*}
where $\bm{K}_d, \bm{K}_p \in \mathbb{R}^{3\times 3}$ are diagonal, positive definite tuning matrices. Then, we can calculate the desired total thrust $f_\text{T, des}$ in body frame by $f_\text{T, des} = \bm{F}_\text{des}^T\cdot \bm{e}_3$.  
Given that the robot can only produce thrust along the $\bm{z}_b$ axis, we align $\bm{z}_{b,\text{des}}$ with $\bm{F}_\text{des}$, and align $\bm{y}_{b,\text{des}}$ to match the desired yaw $\psi_\text{des}$. Therefore, we can calculate the desired attitude $\bm{R}_\text{des}$ as  

\vspace{-2pt}
\begin{equation*}\label{eq:att_des}
\begin{aligned}
\bm{z}_{b,\text{des}} =& \frac{\bm{F}_\text{des}}{||\bm{F}_\text{des}||}\\
\Ddot{\bm{r}}_\psi =& [\cos \psi_\text{des},\  \sin \psi_\text{des},\  0]^T\\
\bm{y}_{b,\text{des}} =& \frac{\bm{z}_{b,\text{des}} \times \Ddot{\bm{r}}_\psi}{||\bm{z}_{b,\text{des}} \times \Ddot{\bm{r}}_\psi||}\\
\bm{R}_\text{des} =& [\bm{y}_{b,\text{des}} \times \bm{z}_{b,\text{des}},\  \bm{y}_{b,\text{des}},\  \bm{z}_{b,\text{des}}]
 \end{aligned}
\vspace{-2pt}
\end{equation*}
where the operator $\times$ denotes the cross product. 
After calculating the desired attitude, we input it into a nonlinear attitude tracking controller to regulate the orientation of the robot. A Quaternion-based controller~\cite{brescianini2013nonlinear} is adopted in this work, but other attitude tracking methods can achieve similar performance. We refer the reader to the PX4 firmware~\cite{meier2015px4} for details about the PID bodyrate control and mixer.   

When collisions occur, the compliant arm compresses with increased estimated contact force ($\hat{f}_e$). Collisions are detected when $\hat{f}_e$ reaches a threshold ($\hat{f}^\ast_{e} = 25$ N). The maximum estimated contact force is measured ($\hat{f}_{e, \text{max}}$), and the collision handling is started when the estimated contact force falls below $\hat{f}^\ast_{e}$ following the detection. We revise the collision handling method in our prior work~\cite{liu2021arq} to generate trajectories to reach a setpoint at a distance proportional to $\hat{f}_{e, \text{max}}$. We use $\bm{r}_c, \Dot{\bm{r}}_c$ to denote the position and velocity of the robot in the inertial frame when the collision handling is started, as well as the rotation matrix ($\bm{R}_c$). The new setpoint ($\bm{r}_n$) in the inertial frame can be written as
\begin{equation}\label{eq:setpoint}
\bm{r}_n = \bm{r}_c -  (\eta \hat{f}_{e, \text{max}} + d_0) \bm{R}_c\bm{x}_b\;,
\vspace{-2pt}
\end{equation}
where $\eta$ and $d_0$ are constants ($\eta = 0.01$ m/N, $d_0 = 0.2$ m).  During collision handling, the robot tracks a smooth (polynomial) trajectory so that for $t\in [t_0, t_T]$, $\bm{r}(t_0) = \bm{r}_c$, $\bm{\Dot{r}}(t_0) = \Dot{\bm{r}}_c$ and it stops at $\bm{r}(t_T) = \bm{r}_n$. The time interval is computed based on maximum accelerations and velocities~\cite{richter2016polynomial}.

\section{Contact-Prioritized Planning}
Contrary to collision-inclusive local re-planners~\cite{lu2022online, zha2021exploiting}, this work proposes an intuitive global planner to exploit impact resilience. We draw motivation from the use case of aerial robots rapidly traversing forest-like environments, whereby maps contain isolated cylindrical obstacles with constant radius ($d_r$). With an intention to utilize the impact resilience, our proposed contact-prioritized (CP) planner prioritizes collisions to facilitate navigation.     

\vspace{-6pt}
\begin{figure}[!h]
\begin{center}
\includegraphics[width=0.225\textwidth]{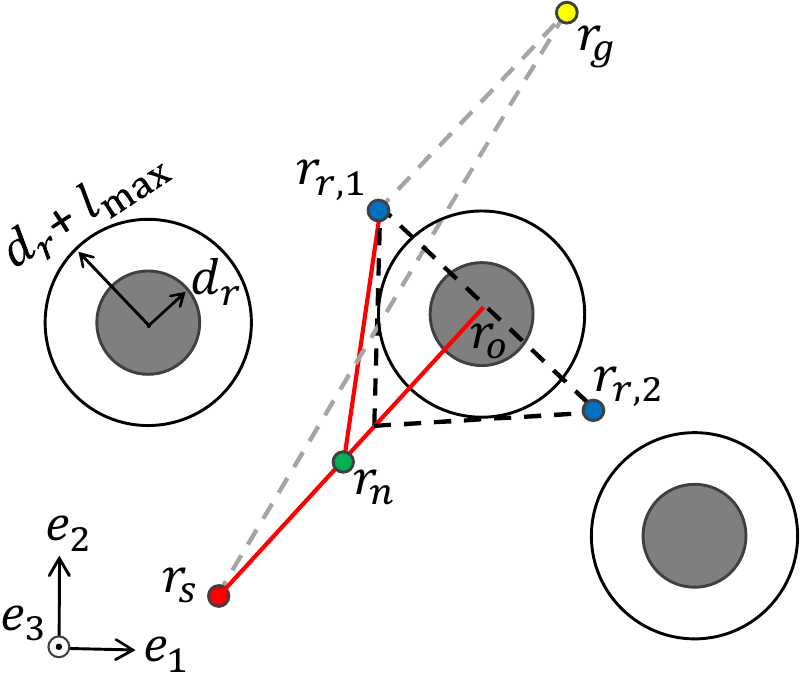}
\end{center}
\vspace{-6pt}
\caption{A novel planning method (CP) to prioritize contact to facilitate navigation in cluttered environments. }
\label{fig:plan}
\vspace{-6pt}
\end{figure}

As shown in Fig.~\ref{fig:plan}, the robot starts at $\bm{r}_s$ while the goal is at $\bm{r}_g$, which share the same altitude ($\bm{e}_3$), thus the navigation is simplified as 2-dimensional (2D) with constant $\bm{e}_3$.\footnote{~Note that the collision recovery has varying altitude but the setpoint $\bm{r}_n$ shares the same $\bm{e}_3$ value.}  The robot is simplified as a ball with a radius of $l_\text{max}$, thus augmented obstacles have a radius of $d_r+l_\text{max}$. The planner starts with drawing a line $\overleftrightarrow{\bm{r}_s \bm{r}_g}$ (gray dashed), and checks if the line intersects with any augmented obstacles. If intersections occur, the robot moves toward the center of the first obstacle $\bm{r_o}$ while controlling yaw to face $\bm{r}_o$ along $\bm{x}_b$ axis, collides and stabilizes at $\bm{r}_n$ as described in \eqref{eq:setpoint}. After recovery, two added waypoints $\bm{r}_{r,\{1,2\}}$ are found, which lie on the line perpendicular to $\overleftrightarrow{\bm{r}_s \bm{r}_o}$ (black dashed) with a distance of $\sqrt{2}(d_r+l_\text{max})$. The robot moves to the added waypoint closer to the goal ($\bm{r}_{r,1}$ in this case), and repeats exploration with a new starting point ($\bm{r}_s = \bm{r}_{r,1}$) until no obstacles are found along the line $\overleftrightarrow{\bm{r}_s \bm{r}_g}$. The robot follows minimum-snap polynomial trajectories with the desired colliding velocities at the center of the in-contact obstacles.

\vspace{-6pt}
\section{Results}
We present results from four experimental tests: force estimation, impact reduction, collision resilience, and planning. A 12-camera VICON motion capture system over WiFi was used for odometry feedback at a rate of $100$ Hz. The feedback is only used to estimate the state of the robot, which can also be achieved by cameras or laser sensors in outdoor environments. Note that we use accelerometer data $\hat{\bm{a}}_b$ of the rigid robot for collision detection. Hence, collisions are detected whenever $|| \bm{R}\hat{\bm{a}}_b+g\bm{e}_3|| \ge 2g$. The rigid robot employs the same collision handling method with a constant maximum force ($\hat{f}_{e, \text{max}}$ = $80$ N) as it cannot directly estimate contact forces.

\vspace{-6pt}
\subsection{Force Estimation}
\begin{table}[!b]
    \caption{Contact Force Estimation Statistics (10 trials).}
    \vspace{-3pt}
    \centering
    \begin{tabular}{c c c c}
         \toprule
        Case $\backslash$ Impact & $30$ N & $40$ N & $50$ N \\
         \midrule
        Static (N) & $30.18\pm 1.08$ \ & $39.76 \pm 1.62$ & $50.61\pm 1.38$\\
         \midrule
        Dynamic (N) & $30.78\pm 3.16$& $40.66 \pm 4.06$ & $52.11 \pm 4.68$\\
        \bottomrule
    \end{tabular}
    \label{tab:estimate}
\end{table}

First, we study the force estimation of the compliant arm. Two cases are considered: when s-ARQ is placed on the ground (static) and while hovering (dynamic). We use a digital force gauge for ground truth. We apply constant forces to s-ARQ at $30, 40, 50$ N in the static case; in the dynamic case we apply impact forces (hit and release) of the same values. Table~\ref{tab:estimate} shows the mean and standard deviation values of ten consecutive trials for each case. Despite sensor noise, results show relatively accurate contact force estimation when the robot is static. The estimation accuracy deteriorates in the dynamic case. This can be associated with vibrations in flight; yet, impact forces also contain larger errors when the robot is flying. Nevertheless, experimental results validate the feasibility of estimating contact forces using the developed compliant arm while in flight.        

\begin{figure}[!t]
\vspace{1pt}
\begin{center}
\includegraphics[trim={0cm 0 1cm 0cm}, width=0.48\textwidth]{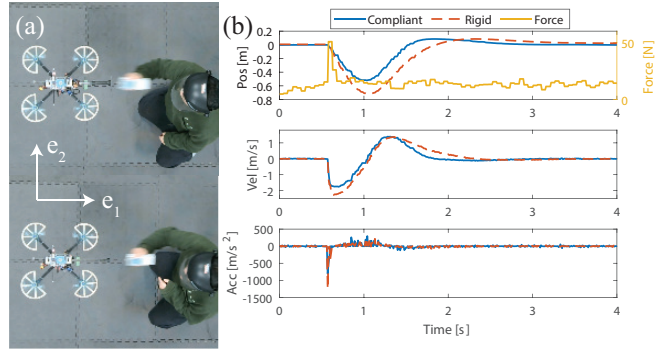}
\end{center}
\vspace{-6pt}
\caption{(a) Snapshots and (b) states tracking of a sample test to evaluate contact force estimation.}
\label{fig:estimate}
\vspace{-12pt}
\end{figure}

Further, we study the effect of the embedded compliance on the response to external impacts. We apply an impact force of $50$ N along the $\bm{e}_1$ axis to both s-ARQ and its rigid counterpart, Quad (Fig.~\ref{fig:estimate}(a) top and bottom panels, respectively). Note that both robots have the same weight in the current and all following tests. Figure~\ref{fig:estimate}(b) depicts the position, velocity, and acceleration along the $\bm{e}_1$ axis of the compliant (blue solid) and rigid (red dashed) robots while the yellow-colored curve shown on the top panel denotes the estimated contact force on the s-ARQ robot. 
Note that accelerations are computed based on velocity data from motion capture in the current and all following tests. 

Results show that s-ARQ can detect a contact force of about $50$ N, as desired. In addition, the compliant robot has fewer changes in all states under impact forcing owing to the embedded compliance. This comparison indicates that existing methods that rely on robot states alone may underestimate impact contact forces when there is embedded compliance.   

\vspace{-6pt}
\subsection{Impact Reduction}

\begin{figure}[!t]
\vspace{6pt}
\begin{center}
\includegraphics[trim={0cm 1cm 1cm 0cm}, width=0.42\textwidth]{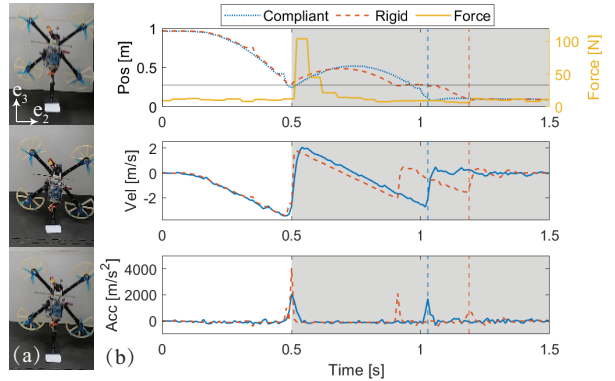}
\end{center}
\vspace{6pt}
\caption{(a) Snapshots and (b) states tracking of a $0.7$ m drop test to study impact reduction. Grey-shaded area denotes states post impact.}
\label{fig:drop}
\vspace{-12pt}
\end{figure}

In the second set of tests we seek to study the impact reduction afforded by the embedded compliance. 
To this end, we employ drop tests. Both robots are fixed vertically ($\bm{x}_b$ facing $-\bm{e}_3$ and $\bm{z}_b$ facing $\bm{e}_1$) before falling to the ground (hard floor mat) from $0.3, 0.5$ and $0.7$ m along $\bm{e}_3$ axis.     


Figure~\ref{fig:drop}(a) shows snapshots from one of the $0.7$ m drop tests for s-ARQ. The compliant arm touches the ground (top), compresses to the minimum length (middle), and then bounces back (bottom).  
Figure~\ref{fig:drop}(b) depicts position, velocity and acceleration tracking along the $\bm{e}_3$ axis of s-ARQ (blue solid) and Quad (red dashed) for a sample $0.7$ m drop test, as well as the estimated contact force (curve in yellow at the top panel). A horizontal black line denotes the $\bm{e}_3$ value ($0.28$ m) when the robot is placed vertically on the ground. 

Results show that both robots have identical position and velocity profiles before touching the ground with a velocity of $-3.46$ m/s. The compliant arm length reduces to its minimum, followed by a saturated force estimation of $104$ N. 
During the impact, s-ARQ has a maximum acceleration of $2,069$ m/s$^2$ while Quad reaches $4,063$ m/s$^2$. Blue and red vertical dashed lines denote that the robot flips to a horizontal state ($\bm{z}_b$ facing $\bm{e}_3$), and therefore lower $\bm{e}_3$ values are observed.          

Further, we repeat ten drop tests at three different $\bm{e}_3$ values for both robots and record the mean and standard deviation of maximum accelerations attained (Tab.~\ref{tab:acc}). Results show that the compliant arm design can help reduce impact by $43.8\%$, $44.3\%$, and $40.3\%$ in the drop tests at $0.3, 0.5$, and $0.7$ m, respectively. These findings demonstrate that our compliant aerial robot design can reduce impact by around $40\%$ with only $4\%$ weight increase.

\begin{table}[!h]
    \caption{Recorded Maximum Acceleration Statistics (10 trials).}
    \vspace{-3pt}
    \centering
    \begin{tabular}{c c c c}
         \toprule
          Robot $\backslash$ Drop Height & $0.3$ m & $0.5$ m & $0.7$ m \\
         \midrule
       Compliant (m/s$^2$) & $1,017\pm 103$ & $1,545\pm 129$ & $2,177 \pm 140$\\
         \midrule
       Rigid (m/s$^2$) & $1,809\pm 186$& $2,774\pm 169$ & $3,649\pm 175$\\
        \bottomrule
    \end{tabular}
    \label{tab:acc}
    \vspace{-12pt}
\end{table}

\vspace{-6pt}
\subsection{Collision Resilience}

We also study the s-ARQ robot's collision resilience using extensive physical collision tests against vertical walls and poles at different velocities and yaw and pitch angles, and compare against the rigid robot, Quad. 
In wall tests, we place a vertical wall at the $\bm{e}_1$ position of $2.45$ m and perpendicular to $\bm{e}_1$ axis. Both robots take off at the $\bm{e}_1$ position of $-1$ m and fly along the $\bm{e}_1$ axis before hitting the wall with zero Euler angles (identity rotation matrix). Owing to the embedded compliance's utility to reduce impact, s-ARQ can sustain wall collisions at a speed of $3.0$ m/s with a $100\%$ success rate for ten consecutive trials (see supplemental video). 
In contrast, the rigid Quad robot can fail at the highest speed collision of $3.0$ m/s because of IMU malfunctions caused by the impact.

\begin{figure}[!t]
\vspace{3pt}
\begin{center}
\includegraphics[trim={0cm 0cm 0cm 0cm}, width=0.49\textwidth]{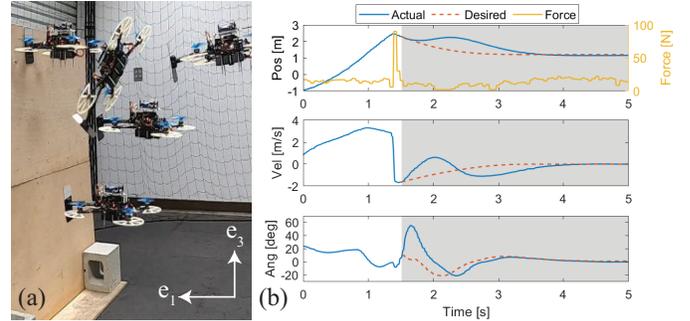}
\end{center}
\vspace{-6pt}
\caption{(a) Snapshots and (b) states tracking of a sample wall collision test at a speed of $3.0$ m/s for s-ARQ.}
\label{fig:3ms}
\vspace{-6pt}
\end{figure}

With reference to Fig.~\ref{fig:3ms}(a), s-ARQ has the compliant arm compressed when colliding with the wall, followed by recovering with large attitude angles and stabilizing at a safe position. Figure~\ref{fig:3ms}(b) depicts the position and velocity of s-ARQ along the $\bm{e}_1$ axis, as well as the pitch angle $\theta$. Blue solid curves denote the actual states while the red dashed ones represent the desired states from the planner. A yellow curve denotes the estimated contact force (top panel), and a grey-shaded area means the recovery control is enabled. 

Results show that s-ARQ touches the wall at a speed of $3.0$ m/s at time $t=1.35$ s. The robot stops and bounces back at a speed of $-1.60$ m/s, during which a maximum contact force of $90$ N is recorded. Recovery control is enabled (desired states) when the estimated force drops below the threshold ($\hat{f}^\ast_{e} = 25$ N) at $t=1.51$ s. The recovery trajectory begins at current position, velocity and attitude states, and stabilizes rapidly at a distance proportional to $\hat{f}_{e,\text{max}}$. The nonlinear controller tracks the aggressive recovery trajectory with large pitch ($\theta_\text{max} = 55^\circ$) for prompt post-collision stabilization. 

\begin{figure}[!t]
\vspace{-12pt}
\begin{center}
\includegraphics[width=0.49\textwidth]{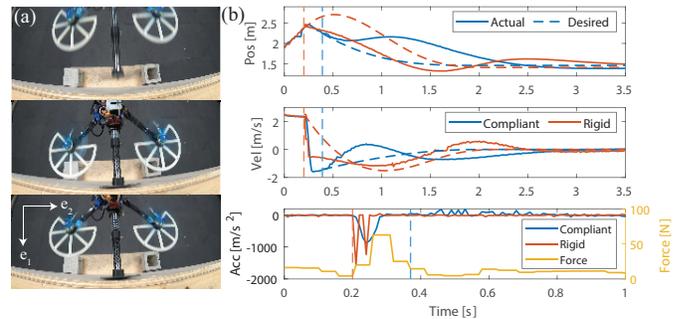}
\end{center}
\vspace{-6pt}
\caption{(a) The compliant arm compressing during a collision. (b) State tracking of both s-ARQ and Quad for $2.5$ m/s wall collisions.}
\label{fig:2dot5}
\vspace{-15pt}
\end{figure}

In addition, we conduct wall collision tests at velocities of $1.5, 2.0$, and $2.5$ m/s along $\bm{e}_1$ axis for both s-ARQ and Quad. Figure~\ref{fig:2dot5}(a) shows a top view of s-ARQ robot wall collision test, where the compliant arm compresses to reduce impact and protect onboard sensors. Figure~\ref{fig:2dot5}(b) visualizes sample trials of wall collision tests at a velocity of $2.5$ m/s for both robots, where solid curves denote measured states while dashed ones represent desired states from the planner. Similarly, blue and red curves visualize position, velocity, and acceleration states of s-ARQ and its rigid counterpart, respectively. Blue and red vertical dashed lines denote the time when recovery controls are enabled for s-ARQ and Quad, and the yellow curve shows the estimated force on s-ARQ.  

Results show that both robots have identical states before the contact at time $t= 0.2$ s, when a collision is detected for the rigid robot. The velocity of Quad changes sharply from $2.5$ to $-0.5$ m/s in $0.05$ s, resulting in a maximum acceleration of $-1,570$ m/s$^2$. Due to the short contact time, the $100$ Hz motion capture feedback causes discontinuity in the velocity tracking. On the contrary, the compliant arm elongates the contact time to $0.1$ s, and reduces the impact to $-876$ m/s$^2$. A maximum contact force of $63$ N is estimated; however, there is a slight delay in the force estimation due to the filter~\eqref{eq:filter}. The recovery control of s-ARQ is enabled at $t=0.37$ s, when the estimated contact force falls below the threshold. s-ARQ collision handling is started $0.17$ s later than Quad's. Still, s-ARQ stabilizes at $t=3.0$ s with a settling time of $2.5$ s, compared to $3.3$ s of Quad. In sum, these findings demonstrate that s-ARQ can stabilize from collisions faster while also mitigating impacts, as compared to the rigid robot. In an effort to demonstrate the preliminary feasibility of our method to apply in practical use cases where high-accuracy localization feedback from motion capture is not available, we experimentally determined that the robot has same success rates when motion capture position feedback was processed (degraded) prior to be sent to the robot in a way that emulates key differences with visual inertial odometry feedback (namely lower accuracy and larger delay).

We further study collisions against pole obstacles, as well as different yaw and pitch collision angles for s-ARQ. The pole obstacle has a radius of $0.15$ m. We drive the s-ARQ robot to have yaw angles $10^\circ$ (left), $0^\circ$ (middle) and $-10^\circ$ (right) collisions against wall and pole obstacles (see supplemental video). Note that the robot has a collision velocity of $2.0$ m/s in all tests. Results show that the robot can stabilize from collisions against walls and poles with different yaw angles. 
Larger angle changes occur during wall collisions in non-zero yaw angles due to the flat geometry of the shield. The collision handling records the current yaw angle at the time of triggering and sustains the angle for stabilization. On the contrary, large angle changes are observed with zero colliding yaw angle in pole collisions, since the robot is not ideally pointing to the geometric center. In addition, s-ARQ can survive wall collisions with yaw angles up to $30^\circ$ (see supplemental video). However, direct contacts between obstacles and propellers occur at larger yaw angles, which pose danger to the robot. 
Further, the robot was experimentally found able to stabilize from large-pitch collisions of $\pm 30 ^\circ$ as well (see supplemental video). 

\vspace{-6pt}
\subsection{Contact-Prioritized (CP) Planning}

In the final set of tests, we study the proposed CP planning method, and compare against A$^\ast$~\cite{likhachev2003ara} and RRT$^\ast$~\cite{karaman2011sampling} in both simulated and physical experiments. The latter help validate the proposed method in practice whereas simulations help better understand the behavior of our CP planning algorithm in terms of its scalability in increasingly cluttered maps, all in relation to standard-of-practice planning algorithm baselines.

\subsubsection{Offline Planning}

\begin{figure}[!t]
\vspace{-9pt}
\begin{center}
\includegraphics[width=0.46\textwidth]{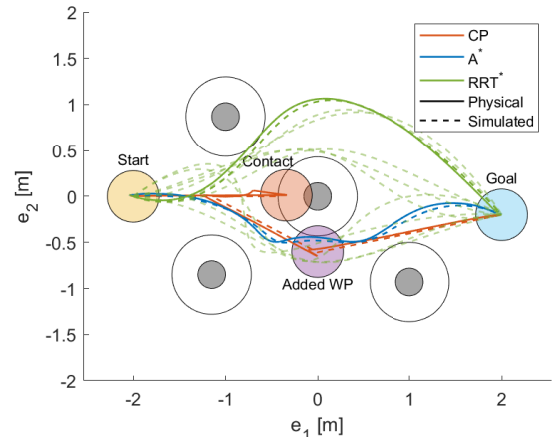}
\end{center}
\vspace{-12pt}
\caption{Physical and simulated trajectories for different planners. }
\label{fig:planning}
\vspace{-6pt}
\end{figure}

The experimental map has a size of $4\times3$ m. Four uniform pole obstacles with a radius of $0.15$ m are located as shown in Fig.~\ref{fig:planning}. The robot starts at $[-2,0]$ and the goal is at $[-2, -0.2]$. We discretize the map with a resolution of $0.1$ m, and run the three algorithms on a Windows machine with an Intel Xeon Processor (3.50 GHz). Four metrics are used to evaluate different methods: planning time, trajectory time, path length and minimum distance to obstacles (clearances). The planning time records the time to find paths in milliseconds, excluding the time for trajectory generation. The trajectory time records the time for robots to reach the goal following the trajectories in seconds. Path lengths denote the Euclidean distance among waypoints (WP). Lastly, we record the clearances between trajectories and augmented obstacles to evaluate the safety against potential collisions as in~\cite{schoels2020ciao}. Note that in-contact obstacles are excluded in the CP planner due to the strong resilience to controlled collisions.

\begin{table}[!h]
    \caption{Comparison Metrics for the Planning Methods.}
    \vspace{-3pt}
    \centering
    \begin{tabular}{c c c c}
         \toprule
         Metrics $\backslash$ Method & \textbf{CP}   & A$^\ast$ & RRT$^\ast$ \\ 
         \midrule
      Plan. Time (ms)  & $2.0$  & $48.9$ & $28.3 \pm 0.9$\\
         \midrule
       Simul. Traj. Time (s)  & $9.4$  & $10.0$ &  $11.3 \pm 1.4$\\
       \midrule
       Phys. Traj. Time (s)  & $9.3$  & $10.1$ & $12.3$\\
       \midrule
       Path Len. (m)  & $5.5$  & $4.2$ & $4.4 \pm 0.2$\\
       \midrule
       Simul. Cl. (m)  & $0.09$  & $0.05$ & $0.01 \pm 0.03$\\
       \midrule
       Phys. Cl. (m)  & $0.10$  & $0.01$ & $-0.02$\\
        \bottomrule
    \end{tabular}
    \label{tab:plan}
    \vspace{-8pt}
\end{table}

Figure~\ref{fig:planning} depicts physical (solid) and simulated (dashed) trajectories of the three planning methods. Note that RRT$^\ast$ results are stochastic with all simulated trials are visualized, however, and only one sample trial is included in the physical testing. Note that polynomials are generated based on waypoints from A$^\ast$ and RRT$^\ast$ planners at a maximum velocity of $1.5$ m/s, while the CP method collides with obstacles at $2.5$ m/s. The robot recovers at position $\bm{r}_n = [-1.25, 0]$ and moves to added WP $\bm{r}_r = [0, -0.61]$ before reaching the goal. 

Comparison metrics of all methods are shown in Tab.~\ref{tab:plan}. It can be observed that simulated results generally match with the physical ones. Results show that our proposed method requires only around 4\% and 7\% planning time compared to A$^\ast$ and RRT$^\ast$ planners. 
In addition, results show that CP leads to the lowest trajectory time. Despite having a
larger path length due to the collision recoveries, the CP planner has almost double obstacle clearances, indicating the enhanced safety of the trajectories. 
This is in fact a benefit of our controlled collision-inclusive planning: by selecting where to collide (safely), the risk for future (unsafe) collisions (as measured by clearances to other obstacles) can be reduced.
                 
We extend planning tests to simulated cluttered maps. The maps have a size of $20\times 20$ m with $30$ pole obstacles (see supplemental video). The obstacles have a uniform radius of $0.3$ m, and are randomly distributed with a clearance of $2.5$ m from center to center. The start is at $[-8,-8]$ while the goal is at $[8,8]$. Ten trials are run for each map and planning method with a discretization resolution of $0.5$ m. Note that $|v_\text{max}|$ of simulated trajectories by A$^\ast$ and RRT$^\ast$ planners were capped at $1.5$ m/s, and a collision speed of $3.0$ m/s was used for trajectories of the CP planner as our prior work indicated higher velocities with collision resilience~\cite{lu2022online}. 

\begin{figure}[!t]
\begin{center}
\includegraphics[width=0.45\textwidth]{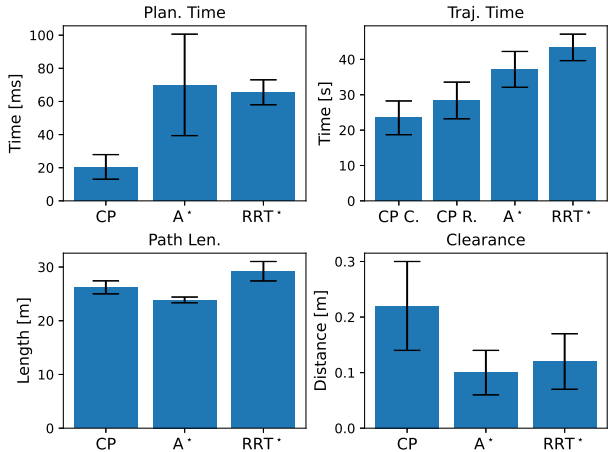}
\end{center}
\vspace{-10pt}
\caption{Comparison metrics for simulated studies on cluttered maps. }
\vspace{-15pt}
\label{fig:simulation}
\end{figure}

Simulated results are listed in Fig.~\ref{fig:simulation} where different planners are evaluated in four metrics as mentioned above. Note that two cases of the CP planner are studied in the trajectory time comparison. The compliant robot (CP C.) has a maximum velocity of $3.0$ m/s and recovery time $2.5$ s, while the rigid robot (CP R.) uses the velocity $2.5$ m/s and time $3.3$ s, as we measured in the collision tests. Results show that the CP planner cost around $30$\% planning time compared to other methods. In the meantime, the results show that the compliant aerial robot with the CP planner saves about $36$\% and $45$\% trajectory time compared to A$^\ast$ and RRT$^\ast$ planners, respectively. The compliant robot saves about $10$\% trajectory time compared to its rigid counterpart under the same planner. On the other, the CP planner has longer path lengths than A$^\ast$ method in the simulation, similar to the observations in the physical test. However, the results show that the trajectories generated by the CP planner have doubled the clearances compared to other methods. To sum up, simulated results help demonstrate that our proposed CP planner can outperform collision-avoidance planning methods A$^\ast$ and RRT$^\ast$ in terms of planning time, trajectory time and path safety, despite longer path lengths than the A$^\ast$ planner.

\vspace{2pt}
\subsubsection{Online Planning}

In support of our method's preliminary feasibility to apply in practical use cases, we also include a simulated study where the robot operates in partially-known maps. The robot can localize obstacles only within a sensing range of $5$ m, which is consistent with practice when the robot relies on cameras or (short-range lightweight and airborne) LiDAR sensors for localization. 
Both CP and A$^\ast$ methods are run online with a re-planning interval of $5$ s. Similarly, ten random maps with 30 obstacles are studied for each planner.

We list the simulated results for both offline and online tasks in Tab.~\ref{tab:horizon}. Note that we also include the trajectory generation time, which stands for the process to convert waypoints to polynomial-based trajectories by solving a constrained optimization problem. Results show that both planning methods have lower planning time in the online task. The A$^\ast$ planing method has a larger decrease, indicating high sensitivity to map size. Results also show that the CP method has lower planning time in both settings. Despite the increased trajectory generation time in partially-known environments, the CP planner costs around $1$\% time of the A$^\ast$ method in generating trajectories, indicating improved efficiency. Both planners have longer trajectory time in the online task due to the limited knowledge of the map. Still, trajectories generated by the CP planner save about $23$\% of the time compared to A$^\ast$ trajectories, despite longer path lengths in both settings. 

\begin{table}[!t]
\vspace{6pt}
    \caption{Comparison Metrics for Different Environments.}
    \renewcommand*{\arraystretch}{1.4}
    \vspace{-3pt}
    \centering
    \begin{tabular}{ c c c c c }
         \toprule
          & Metrics & Offline & Online & Units\\
         \midrule
      \multirow{ 4}{*}{ \textbf{CP} } & Plan. Time  & $20.50 \pm 6.43$  & $16.78 \pm 4.41$ &ms \\
            \cline{2-5}
                & Traj. Gen. Time & $0.079 \pm 0.026$  & $0.318 \pm 0.122$ &  s \\
                \cline{2-5}
         & Traj. Time   & $23.48 \pm 4.77$  & $31.04 \pm 5.27$ & s\\
         \cline{2-5}
       &Path Len. & $26.22 \pm 1.21$  & $28.64 \pm 2.98$ & m\\
       \cline{1-5}
      \multirow{ 4}{*}{A$^\ast$}  & Plan. Time  & $70.02 \pm 30.61$  & $38.51 \pm 4.52$ & ms\\
      \cline{2-5}
              &Traj. Gen. Time   & $32.80 \pm 2.65 $  & $26.17 \pm 7.40$ & s\\
              \cline{2-5}
        &Traj. Time & $37.17 \pm 5.07$  & $40.30 \pm 4.37$ & s\\
        \cline{2-5}
        & Path Len. & $23.88 \pm 0.53$  & $24.25 \pm 1.22$ & m\\
        \bottomrule
    \end{tabular}
    \label{tab:horizon}
    \vspace{-15pt}
\end{table}

\vspace{-2pt}
\section{Conclusion}
In this work, we designed a lightweight compliant arm to sense contacts and reduce collision impact. 
Equipped with the integrated arm, we developed a novel impact-resilient aerial robot, named s-ARQ, to stabilize from high-speed collisions. 
Experimental results show that the compliant robot has only a $4\%$ weight increase but around $40\%$ impact reduction compared to a rigid counterpart. Further, when equipped with a real-time contact force estimator and a non-linear motion controller, the compliant robot can handle collisions while attempting aggressive maneuvers, and stabilize from high-speed wall collisions at $3.0$ m/s with a success rate of $100\%$. 
This impact resilience was also verified with pole obstacle collisions, as well as with different yaw and pitch angles. Our robot is found to result in better performance over both maximum collision speeds and success rates compared to other state-of-the-art methods reported in the literature.

We also proposed and validated in both simulated and physical experiments a planning method for impact-resilient robots that prioritizes contacts to facilitate navigation. 
Physical tests and extended simulations demonstrate that our proposed compliant robot and contact-prioritized planning method can accelerate the computation while achieving shorter trajectory time by relaxing velocity constraints. Despite having a larger path length due to the collision and follow-on recovery, the CP planner leads to higher clearances, indicating enhanced safety. Online planning tests in partially-known environments were also studied. Simulated results further validated the efficiency of the proposed CP planner, with reduced planning and trajectory generation time, shorter trajectory time and increased clearances. Admittedly, the CP planner has longer trajectory time compared to collision-avoidance planning methods when applying velocities constraints. However, the significant saving in computational time and increased trajectory safety may outweigh the increasing path length limitation. The proposed CP planner thus provides positive results to study how to utilize contacts to facilitate navigation, especially when computational time is of essence.     

There are multiple interesting directions to explore in future work. Herein we assumed regular wall and pole obstacles; we plan to extend results to irregular obstacles in 3D space. While collision handling involves basic motion control, it is possible to study force-control-based recovery methods on the compliant robot such as impedance and admittance control~\cite{tomic2017external}. Further, we plan to incorporate camera or laser distance sensors for odometry feedback and deploy impact-resilient aerial robots in outdoor environments. 
Lastly, we adopted a conservative velocity for traversing cluttered environments; studying collision-inclusive high-speed flight is also another direction of future research.


\vspace{-8pt}
\bibliographystyle{IEEEtran}
\bibliography{IEEEabrv, mybib}

\end{document}